\documentclass[conference]{IEEEtran}
\usepackage{cite}
\usepackage{amsmath,amssymb,amsfonts,graphicx}
\usepackage{algorithmic}
\usepackage{graphicx}
\usepackage{textcomp}
\usepackage{xcolor}
\usepackage[section]{placeins}
\usepackage{subfig}
\graphicspath{{imgs/}}

\def\BibTeX{{\rm B\kern-.05em{\sc i\kern-.025em b}\kern-.08em
		T\kern-.1667em\lower.7ex\hbox{E}\kern-.125emX}}
	
\makeatletter
\newcommand{\linebreakand}{%
\end{@IEEEauthorhalign}
\hfill\mbox{}\par
\mbox{}\hfill\begin{@IEEEauthorhalign}
}
\makeatother
	
\begin{document}
	
	\title{Generatively Augmented Neural Network Watchdog for Image Classification Networks}

	\author{\IEEEauthorblockN{Justin M. Bui}
		\IEEEauthorblockA{\textit{Department of Electrical and Computer Engineering}\\
			\textit{Baylor University} \\
			Waco, TX \\
			Justin\_Bui@baylor.edu}
		\and
		\IEEEauthorblockN{Glauco A. Amigo}
		\IEEEauthorblockA{\textit{Department of Electrical and Computer Engineering}\\
			\textit{Baylor University} \\
			Waco, TX \\
			Glauco\_Amigo1@baylor.edu}
		\linebreakand
		\IEEEauthorblockN{Robert J. Marks II, PhD}
		\IEEEauthorblockA{\textit{Department of Electrical and Computer Engineering} \\
			\textit{Baylor University}\\
			Waco, TX \\
			Robert\_Marks@baylor.edu}
	}
	
	\maketitle
	\begin{abstract}
	The identification of out-of-distribution data is vital to the deployment of classification networks. For example, a generic neural network that has been trained to differentiate between images of dogs and cats can only classify an input as either a dog or a cat. If a picture of a car or a kumquat were to be supplied to this classifier, the result would still be either a dog or a cat. In order to mitigate this, techniques such as the neural network watchdog have been developed. The compression of the image input into the latent layer of the autoencoder defines the region of in-distribution in the image space. This in-distribution set of input data has a corresponding boundary in the image space. The watchdog assesses whether inputs are in inside or outside this boundary. This paper demonstrates how to sharpen this boundary using generative network training data augmentation thereby bettering the discrimination and overall performance of the watchdog.
	\end{abstract}

	\section{Introduction}	
	Neural networks and the techniques associated with machine learning are readily being used in a variety of applications. In fact, machine learning has become ubiquitous throughout many industries, including the medical, automotive, and energy industries \cite{Garg,Hanga,Ozbayoglu,Paleyes,Rabe,Sharp}. Furthermore, the development of such techniques has been spurred by the continuous development and expanded capabilities of tools such as PyTorch and TensorFlow \cite{Abadi,Paszke}. 
	
	An increasing need for quality training data goes hand in hand with the development of new machine learning techniques. Data augmentation and generation have become a standard practice in the development of deep neural networks \cite{Taylor,Miko,Perez,Shorten,Wen}. As systems become more complex, and pipelines are transitioned from evaluation to production environments, the need for online out-of-distribution detection increases \cite{Haug}. One technique to help mitigate the impact of out-of-distribution inputs is the neural network watchdog. This paper describes the design and development process for a multi-tiered neural network watchdog  designed to improve the on-line performance of an image classification neural network. This is achieved by a more effective sharpening of the boundary of the image space, resulting in more accurate discrimination of the in-distribution from out-of-distribution data as dictated by the latent layer of the autoencoder.

	\section{Background}
	One of the challenges associated with the development of classification neural networks is the detection and classification of out-of-distribution data. Abbasi et al. \cite{Abbasi} provide an excellent introduction to the issues and metrics associated with out-of-distribution data in classification networks. Out-of-distribution detection is often times closely coupled to anomaly detection \cite{Golan,Hendrycks,Nguyen} (a.k.a. novelty filtering \cite{El-Sharkawi, Guttormsson, Streifel,Thompson2002}.) Numerous anomaly detection techniques have been proposed \cite{Wei,Pradhan,Sabokrou} which demonstrate some capabilities of detecting out-of-distribution data. One of the more interesting techniques for anomaly detection is the use of autoencoders \cite{Marchi,Qi,Sakurada,Zhou}. Autoencoders have demonstrated an interesting array of capabilities, such as implicit learning \cite{Thompson2002}, missing sensor restoration,
	\cite{Narayanan,Thompson2003}, data denoising \cite{Gondara,Lu} and image enhancement \cite{Kim,Lore,Park}.   
 
	The neural network watchdog \cite{Bui}, is a technique for out-of-distribution detection. Initially, an autoencoder is used to learn latent representations of in-distribution data, and then reproduce the original input based on the representation. As the autoencoder is trained on in-distribution data, the representations produced by out-of-distribution data result in larger error during reconstruction. The calculated error is then compared to a threshold based on the desired level of detection. This threshold operation acts as an input filter to the original classification or regression network. The evaluation and selection of an appropriate error function depends on the input data type and target system performance. For example, a time-series dataset may use mean bias error, root mean squared error or unscaled mean bounded relative absolute error \cite{Chen} as a metric, whereas image based data may use structural similarity techniques \cite{Wang,Wang2}, univariate image difference, or change vector analysis \cite{Bruzzone}. Additional error determination methods have been discussed \cite{Chalom,Johnson}. 
	
	Preliminary results have demonstrated successful implementations of neural network watchdogs \cite{Bui,Bui2,Fredieu}. The watchdog has been augmented to a multi-tiered watchdog \cite{Galan}. The multi-tiered watchdog uses several different out-of-distribution detection mechanisms designed in coordination to mitigate some of the performance tradeoffs associated with watchdog development.  
	
	While leveraging multiple detection mechanisms, the watchdog concept can be further improved using data augmentation techniques. The watchdog relies on the assessment of the boundary of the latent space in the autoencoder. 
	
	Without using an autoencoder, classification boundary points can be found directly from the trained neural network classifier using neural network inversion \cite{Cohn}, \cite{Duren}, \cite{Hwang1990A,Hwang1990B,Jensen1997,Jensen1999A,Jensen1999B,Kassabalidis} , \cite{Oh,Reed,Tsang1991,Tsang1992}. Here, a boundary point location is specified at the output of a trained neural network and the corresponding input is found.
 
	Data augmentation has moved beyond neural network inversion techniques with the development of GANs \cite{Goodfellow}. Training data can now be generated using a variety of GANs \cite{Karras,Mao,Metz}. Papers presented by Creswell et al. \cite{Creswell} and Goodfellow et al. \cite{Goodfellow2} provide excellent reviews of the capabilities of GANs. Work done by Zhu et al. and previous work demonstrates the use of GANs in the development of augmented training data \cite{Zhu,Galan}. 
	
	\section{The Neural Network Watchdog}
\begin{figure*}[h!]
		\begin{center}
			\includegraphics[width=.8\textwidth]{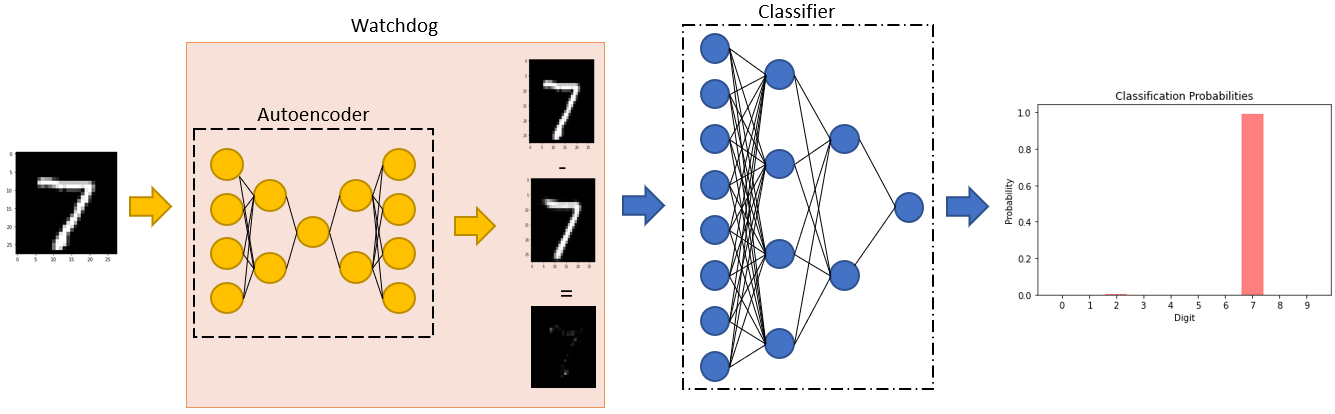}
			\caption{An example of image data flowing through a watchdog guarded image classifier.}
			\label{SWD}
		\end{center}
	\end{figure*}

	The neural network watchdog has been successfully demonstrated as a method to improve classifier performance with mixed-distribution input data \cite{Bui,Bui2,Fredieu}. The watchdog first uses a generative component to reconstruct input data, such as an autoencoder. The reconstruction is then compared to the original input data using an error function. An application-specific criterion is then used to determine input validity, and whether or not the original input will be passed into the classifier. Figure~\ref{SWD} shows an example of how data flows through a watchdog guarded image classification network.   
	
	\subsection{Network Structures}
\begin{figure*}[h!]  
		\begin{center}
			\includegraphics[width=.8\textwidth]{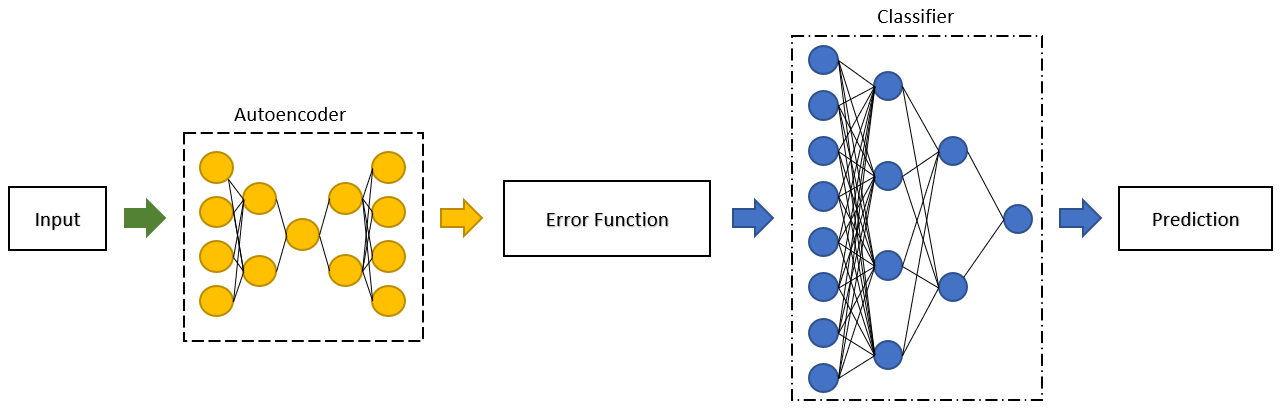}
			\caption{The general structure of a  Disjointed Watchdog.   }
			\label{DJW}
		\end{center}
	\end{figure*}

\begin{figure*}[!t]
		\begin{center}
			\includegraphics[width=.65\textwidth]{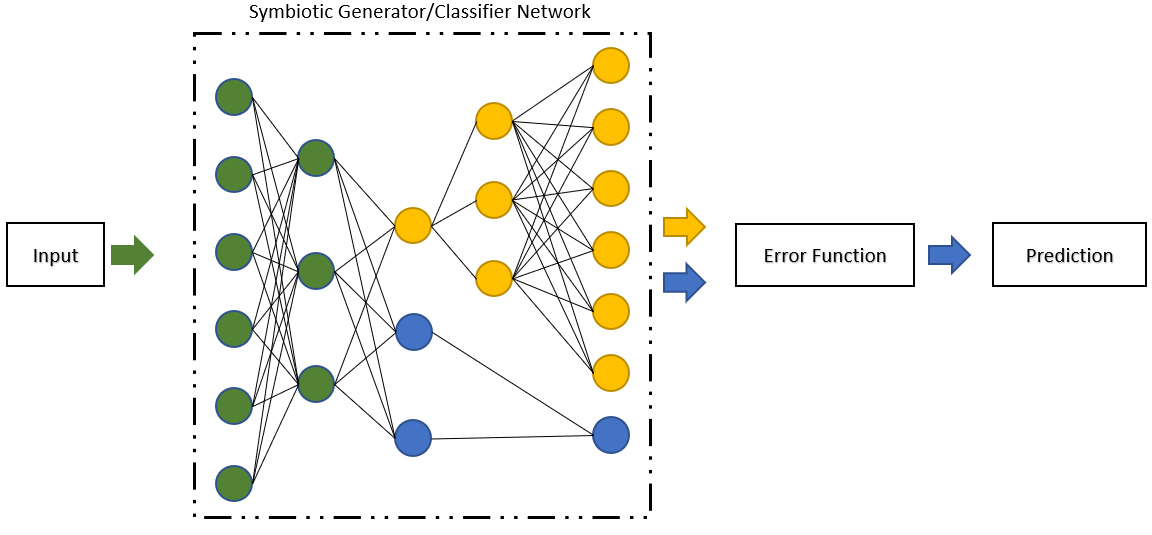}
			\caption{The general structure of a Symbiotic Watchdog.}
			\label{SBW}
		\end{center}
	\end{figure*}

\begin{figure*}[!t]
		\begin{center}
			\includegraphics[width=.85\textwidth]{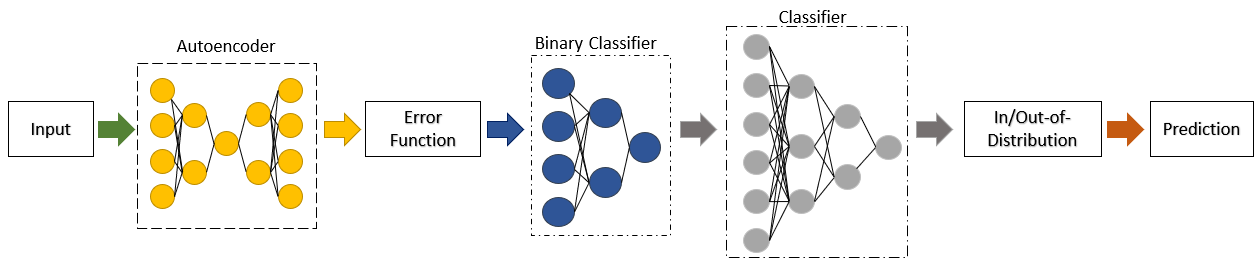}
			\caption{The general structure of a Multi-tiered Watchdog Structure}
			\label{CWD}
		\end{center}
	\end{figure*}

	The neural network watchdog can be implemented in a variety of architectures, including disjointed, symbiotic, and multi-tiered. Examples of these architectures can be seen in Figures~\ref{DJW}, \ref{SBW}, and \ref{CWD}. Discussions about the differences and trade offs between disjointed, symbiotic and multi-tiered architectures can be found in previous work \cite{Galan,Bui2}.
	
	The disjointed watchdog (see Fig. \ref{DJW}) consists of a regenerative autoencoder and a primary classifying neural network, which are trained independently using the same training data \cite{Bui}. The selection of an error function is based on the desired system performance.

	The symbiotic watchdog (see Fig. \ref{SBW}) consists of a single-input multi-output hybrid generator/classifier network \cite{Bui2}. The network is trained on a single dataset, with bias towards classifier or generator as necessary. As seen in Figure~\ref{SBW}, both the generated output and classification are produced simultaneously. In order to determine input validity, the generated reconstruction is provided to an error function, and the classification output is only used when the generated reconstruction meets the determined error criteria.
	
	The multi-tiered watchdog (see Fig. \ref{CWD}) is comprised of multiple layers of outlier detection techniques, such as a combined autoencoder and binary classifier \cite{Galan}. With multiple watchdog layers, layers can be individually tuned in order to achieve the desired results. 
	
	\subsection{Watchdog Threshold Functions} 
	Fundamental to the successful implementation of a neural network watchdog is the selection of an appropriate threshold function. Previous work demonstrates the use of a root mean squared error based threshold function \cite{Galan,Bui,Bui2}.

	\subsection{Generative Augmentation}
	Data augmentation is a common practice to help improve the accuracy of a neural network. Data augmentation can occur in a variety of ways, such as image manipulation and data generation.	Previous work \cite{Galan} uses a GAN to generate training data to be used with a binary classification layer. The additional training data can be used a number of ways:
	
	\begin{enumerate}
		\item Increased amount of training examples for unbalanced datasets.
		\item Training a binary classifier for in and out of distribution detection.
		\item Create a ${(N+1)}$th class to an existing classifier for unrecognized inputs.
	\end{enumerate}

	\section{Results}
	This paper proposes a multi-tiered image classification watchdog for outlier detection and removal. The multi-tiered watchdog is added to the input data pipeline, which in turn feeds into the core classification network. By adding the watchdog to the data pipeline, out-of-distribution images are identified and discarded prior to presentation to the classification network. Doing so decreases computational load and improved mixed-distribution classification accuracy.

	\subsection{Developing a Multi-tiered Watchdog}
	A multi-tiered watchdog consists of multiple outlier detection techniques that are combined sequentially.  The first layer is comprised of an autoencoder and an error function. The second layer is a binary classifier to determine whether input data is in or out of distribution. The multi-tiered watchdog output is then fed in to a traditional multi-class convolutional neural network classifier. 
	
	\subsection{Training and Evaluation Datasets}
\begin{figure}[h!]
		\begin{center}
			\includegraphics[width=.4\textwidth]{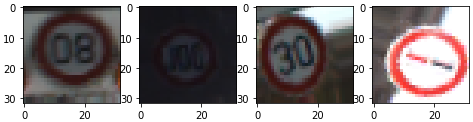}
			\caption{Examples from the in-distribution German traffic sign dataset.}
			\label{GTSRB}
		\end{center}
	\end{figure}

	To evaluate the efficacy of the multi-tiered watchdog, a combination of datasets will be used to train and evaluate the networks. The German Traffic Sign Benchmark (\emph{GTSRB}) dataset \cite{Stallkamp} is used as the primary dataset, and is considered to be in-distribution. Examples of the GTSRB dataset, images  of 43 different traffic signs, can be seen in Figure~\ref{GTSRB}. Approximately 39,000 images are used to train the watchdog and core classifier with an additional 12,600 images to be used to independently evaluate all of the networks.
	
\begin{figure}[h!]
		\begin{center}
			\includegraphics[width=.4\textwidth]{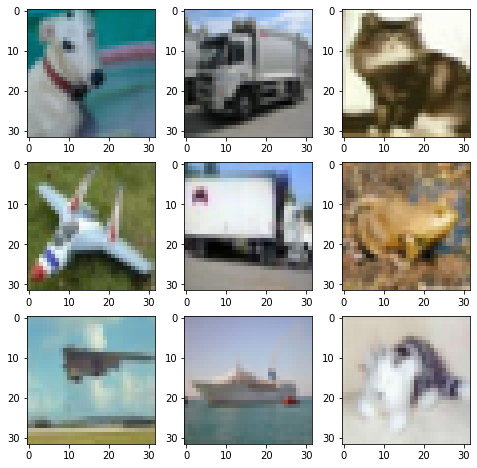}
			\caption{Example images from the CIFAR-10 and CIFAR-100 datasets, which are added to create a mixed-distribution evaluation dataset.}
			\label{cifar}
		\end{center}
	\end{figure}

	To augment the evaluation dataset, 10,000 additional images will be used from 
	each the CIFAR-10 and CIFAR-100 datasets to create a 32,600 image mixed distribution dataset. In order to effectively evaluate the networks two series of labels are used for this dataset. The first is the original image label, and the second is a binary label corresponding to either in or out of distribution. These series of labels will allow for an effective evaluation of the core and binary classifiers. Examples of the CIFAR-10 and CIFAR-100 images can be seen in Figure~\ref{cifar}.

	\subsection{Training Process}
	In order to successfully train the multi-tiered watchdog, the training process is divided up into four sub-processes:
	
	\begin{enumerate}
		\item Train the autoencoder.
		\item Define generative network parameters.
		\item Train the binary classifier.
		\item Train the core classifier.
	\end{enumerate} 

	The core classifier and autoencoder can be trained independently. The generator and binary classifier are dependent on the performance of the autoencoder, and therefore must be trained after. Both the autoencoder and the core classifier will be trained on the in-distribution GTSRB images. The binary classifier is then trained on a combination of in-distribution GTSRB images, as well as near boundary images created by the generator \cite{Galan}. 
	
	\subsubsection{Training the Autoencoder}
	The autoencoder is trained on in-distribution data, with the intent to regenerate the input with minimal mean squared error. Unlike some datasets, such as the MNIST or CIFAR datasets, the GTSRB dataset has relatively low numbers of images per each class. To compensate for this fact, the training data is augmented using tools such as TensorFlow's built in Image or Keras APIs. Such tools allow for traditional image and computer vision processing techniques, including image rotation, cropping, flip, and grayscale conversion. This augmentation allows the autoencoder to produce higher quality results on relatively small training sets. 

	\subsubsection{Training the Binary Classifier}
	A key component of training the binary classifier is the generation of new data. A generator is designed to create data which closely resembles near-threshold conditions \cite{Galan}. The near threshold data is classified as out-of-distribution for training purposes. Once the parameters have been defined and additional training data has been generated, the binary classifier is trained to differentiate between in-distribution and generated out-of-distribution data. One important note is that as adjustments are made to the autoencoder, they must also be made to the generator and binary classifier.

	\subsubsection{Training the Core Classifier}
	The process for training the core classifier is similar to training the autoencoder. Training data is augmented, as previously mentioned, through the use of the TensorFlow APIs, and the classifier is trained using categorical cross entropy training loss. 
	
\begin{figure}[h!]
		\begin{center}
			\includegraphics[width=.5\textwidth]{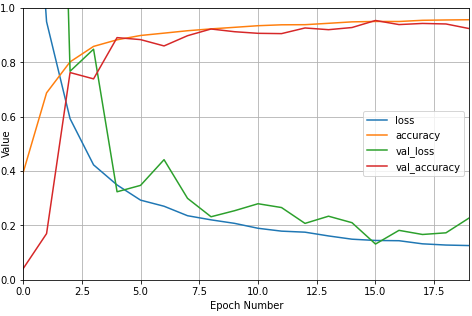}
			\caption{Classifier training history for validation and training accuracy and losses. The validation dataset is a subset of the training data, and does not overlap with the  data that will be used for evaluation.}
			\label{classhist}
		\end{center}
	\end{figure}

	Figure~\ref{classhist} shows the training accuracy and losses for the classifier.

	\section{Evaluating The Networks}
	Once all of the networks have been trained, they are evaluated using the mixed-distribution dataset. The individual performance of the core classifier, autoencoder and binary classifier are first individually evaluated. The autoencoder and binary classifier are then added to the data input pipeline, and the combined system is  evaluated.
	
	\subsection{Unguarded Core Classifier Performance}

\begin{figure}[h]
		\begin{center}
			\includegraphics[width=.5\textwidth]{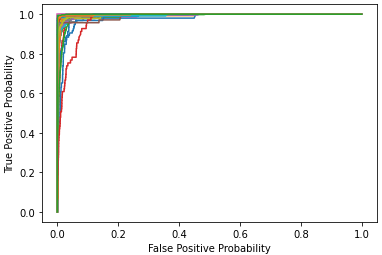}
			\caption{The ROC curves of the core classifier performance evaluated on the 43 independent classes of the \emph{GTSRB} dataset.}
			\label{classall}
		\end{center}
	\end{figure}

\begin{figure}[h]
		\begin{center}
			\includegraphics[width=.5\textwidth]{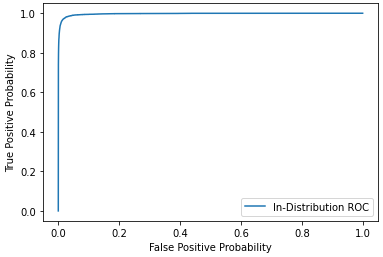}
			\caption{The normalized ROC curve of the core classifier performance on the \emph{GTSRB} dataset.}
			\label{classnorm}
		\end{center}
	\end{figure}

	The performance of the core classifier is evaluated using both the German traffic sign and mixed-distribution datasets. The isolated performance of the classifier, as shown in Figure~\ref{classall}, is determined by evaluation using the in-distribution data, and serves as the performance baseline. The normalized performance of the core classifier can be seen in Figure~\ref{classnorm}.
	
\begin{figure}[h!]
		\begin{center}
			\includegraphics[width=.48\textwidth]{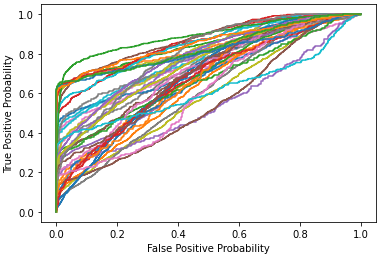}
			\caption{The ROC curves of the core classifier performance evaluated on the mixed distribution dataset.}
			\label{unguardall}
		\end{center}
	\end{figure}

\begin{figure}[h!]
		\begin{center}
			\includegraphics[width=.48\textwidth]{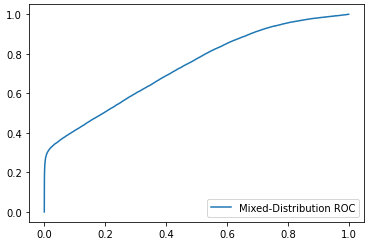}
			\caption{The normalized ROC curve of core classifier performance evaluated on the mixed distribution dataset.}
			\label{unguardnorm}
		\end{center}
	\end{figure}

	In order to understand the impact of out-of-distribution data, the same evaluation is performed using the mixed-distribution dataset. The online performance of the core classifier, as shown in Figures~\ref{unguardall}  and \ref{unguardnorm}, demonstrates the reduced accuracy due to out-of-distribution data. With a ratio of ${1/3}$ in-distribution to ${2/3}$ out-of-distribution data, our normalized accuracy and precision drops to approximately 33\%, as expected. 
	
	\subsection{Autoencoder Performance and Threshold Selection}
\begin{figure}[h!]
		\begin{center}
			\includegraphics[width=.45\textwidth]{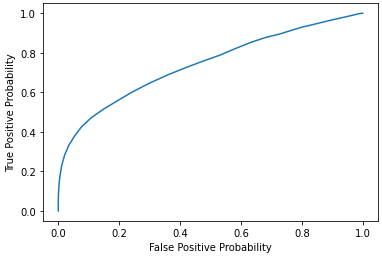}
			\caption{The ROC curve for the autoencoder's performance, as a function of structural similarity score.}
			\label{ae_roc}
		\end{center}
	\end{figure}

	The performance of the autoencoder is evaluated by measuring the structural similarity score, defined in Equation~\ref{eq:SSIM}, of the original and regenerated images and comparing it to the determined structural similarity threshold value. Figure~\ref{ae_roc} shows the ROC curve for binary classification of in-distribution vs out-of-distributed as across several structural similarity thresholds. 
	
	\begin{equation}
		SSIM(x,y) = \frac{(2\mu_x\mu_y + C_1) + (2 \sigma _{xy} + C_2)} 
		{(\mu_x^2 + \mu_y^2+C_1) (\sigma_x^2 + \sigma_y^2+C_2)}
		\label{eq:SSIM}
	\end{equation}	
	The constants $C_1$ and $C_2$ are defined by Equations~\ref{eq:C1} and \ref{eq:C2}, where L represents the data input range value, and $K_1$ and $K_2$ are constants $<<$ 1 \cite{Wang}.
	
	\begin{equation}
		C_1 = (K_1*L)^2
		\label{eq:C1}
	\end{equation}
	\begin{equation}
		C_2=(K_2*L)^2
		\label{eq:C2}
	\end{equation}

\begin{figure}[h!]
		\begin{center}
			\includegraphics[width=.45\textwidth]{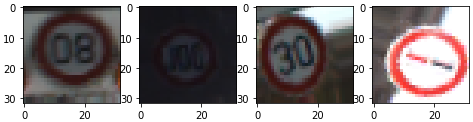}
			\caption{The original \emph{GTSRB} sign images.}
			\label{germ_orig}
		\end{center}
	\end{figure}

\begin{figure}[h!]
		\begin{center}
			\includegraphics[width=.45\textwidth]{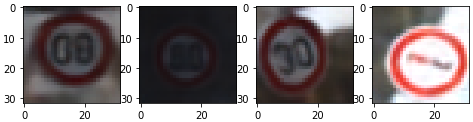}
			\caption{An example of the autoencoder regenerated \emph{GTSRB} sign images.}
			\label{germ_regen}
		\end{center}
	\end{figure}

\begin{figure}[h!]
		\begin{center}
			\includegraphics[width=.45\textwidth]{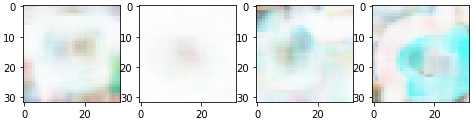}
			\caption{The structural similarity images of the \emph{GTSRB} examples.}
			\label{germ_ssim}
		\end{center}
	\end{figure}

	Examples of the autoencoder image regeneration can be seen in Figure~\ref{germ_regen}, along with their original images in Figure~\ref{germ_orig} and the structural similarity difference in Figure~\ref{germ_ssim}. The structural similarity scores for these images are found in Table~\ref{tab1}.

\begin{figure}[h!]
		\begin{center}
			\includegraphics[width=.45\textwidth]{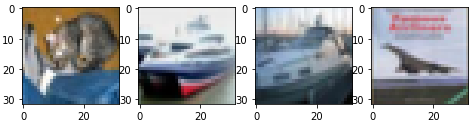}
			\caption{The original CIFAR images.}
			\label{cf_orig}
		\end{center}
	\end{figure}

\begin{figure}[h!]
		\begin{center}
			\includegraphics[width=.45\textwidth]{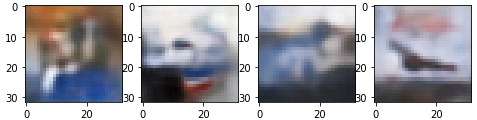}
			\caption{An example of the autoencoder regenerated CIFAR images.}
			\label{cf_regen}
		\end{center}
	\end{figure}

\begin{figure}[h!]
		\begin{center}
			\includegraphics[width=.45\textwidth]{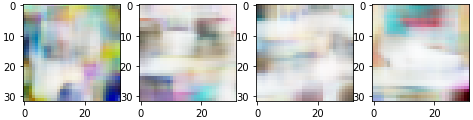}
			\caption{The structural similarity images of the CIFAR examples.}
			\label{cf_ssim}
		\end{center}
	\end{figure}

		As a reference, the autoencoder regeneration of a sample of CIFAR images is provided in Figure~\ref{cf_regen}, as well as the respective original images in Figure~\ref{cf_orig} and the difference in Figure~\ref{cf_ssim}.
	
	\begin{table}[h!]\centering
		\caption{Structural Similarity Scores}\label{tab1}
		\begin{tabular}{|l|l|l|l|l|}
			\hline
			 & Image 1 & Image 2 & Image 3 & Image 4 \\ 
			\hline
			Signs Scores: & 0.9303 & 0.9715 & 0.9184 & 0.8903 \\ 
			\hline
			CIFAR Scores: & 0.7488 & 0.7767 & 0.8624 & 0.8318 \\ 
			\hline
		\end{tabular}
	\end{table}

	Based on these results, and depending on the desired level of protection, a structural similarity score of between 0.83 and 0.87 is an appropriate selection. For the remainder of this proof of concept, a structural similarity score of 0.85 is used.
	
	\subsection{Binary Classifier Data Generation}
\begin{figure}[h!]
		\begin{center}
			\includegraphics[width=.45\textwidth]{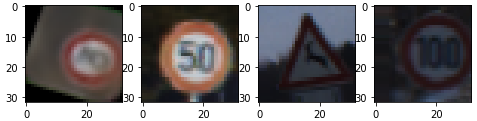}
			\caption{Generated binary class images with a target structural similarity score of 0.90}
			\label{genimgs}
		\end{center}
	\end{figure}

	With the baseline autoencoder trained and the error function established, a generative network is used to create additional training and evaluation data. The newly generated data is used to train the binary classifier. The generative network is designed to generate images with a target structural similarity score. Based on the previously selected 0.85 score, a target score of 0.90 will be used. Example generated images may be seen in Figure~\ref{genimgs}.
	
	\subsection{Binary Classifier Performance}
\begin{figure}[h!]
		\begin{center}
			\includegraphics[width=.45\textwidth]{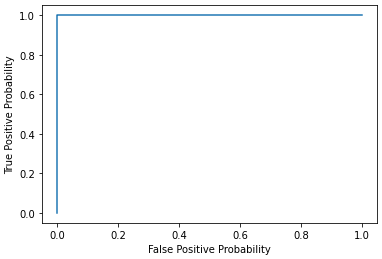}
			\caption{The ROC curve for the binary classifier, evaluated on the generated test dataset.}
			\label{bcroc}
		\end{center}
	\end{figure}

	The performance of the binary classifier is evaluated on an independent subset of the generated data. The goal of the binary classifier is to differentiate between in- and out-of-distribution images that reside close to the dataset manifold which have been permitted by the autoencoder layer. The ROC curve for the binary classifier can be seen in Figure~\ref{bcroc}.
	
	\subsection{Final performance of the Guarded Classifier}
\begin{figure}[h!]
		\begin{center}
			\includegraphics[width=.45\textwidth]{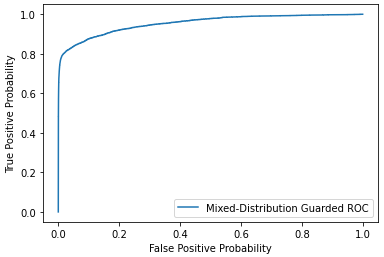}
			\caption{The normalized ROC curve for the watchdog guarded image classifier.}
			\label{final_roc}
		\end{center}
	\end{figure}

	The efficacy of the multi-tiered watchdog is determined by the combined performance of the autoencoder and the binary classifier. With the watchdog fully implemented, the final system performance is evaluated. The ROC curve seen in Figure~\ref{final_roc} shows the watchdog guarded performance on the mixed-distribution evaluation dataset.

\begin{figure}[h!]
		\begin{center}
			\includegraphics[width=.45\textwidth]{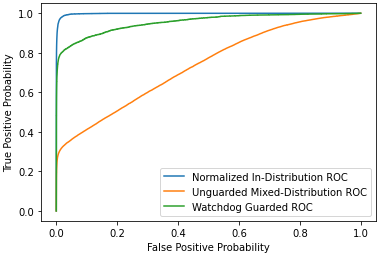}
			\caption{The normalized ROC curves for the unguarded, watchdog guarded, and ideal core classifier.}
			\label{final_all}
		\end{center}
	\end{figure}

	A final comparison of the ROC curves for the unguarded, guarded, and in-distribution only datasets may be seen in Figure~\ref{final_all}. The introduction of the watchdog results in significantly improved performance when compared to the unguarded core classifier alone.
		
	\section{Conclusion}
	A multi-tiered neural network watchdog is developed to improve online classification neural network performance with respect to out-of-distribution data. A generative network is trained to produce data which resembles out-of-distribution data permitted by the autoencoder. A binary classifier is trained to differentiate between the permitted out-of-distribution data and in-distribution data. The combination of the autoencoder and binary classifier results in reliable out-of-distribution detection and filtering. The final results presented demonstrate classification improvements  when evaluated on mixed-distribution input data.


\begin{thebibliography}{00}
	
	
	\bibitem{Abadi} Abadi, Martín, et al. "Tensorflow: A system for large-scale machine learning." 12th {USENIX} symposium on operating systems design and implementation ({OSDI} 16). 2016.
	
	\bibitem{Abbasi} Abbasi, Mahdieh, et al. "Toward Metrics for Differentiating Out-of-Distribution Sets." arXiv preprint arXiv:1910.08650 (2019).
	
	\bibitem{Galan} Amigo Galán, Glauco A., Justin Bui, and Robert J. Marks. "Cascade Watchdog: A Multi-tiered Adversarial Guard for Outlier Detection." arXiv preprint arXiv:2108.09375 (2021).
	
	\bibitem{Bruzzone} Bruzzone, Lorenzo, and Diego F. Prieto. "Automatic analysis of the difference image for unsupervised change detection." IEEE Transactions on Geoscience and Remote sensing 38.3 (2000): 1171-1182.
	
	\bibitem{Bui} Bui, Justin, and Robert J. Marks II. ``Autoencoder Watchdog Outlier Detection for Classifiers." arXiv preprint arXiv:2010.12754 (2020).
	 
	\bibitem{Bui2} Bui, Justin, and Robert J. Marks II. "Symbiotic Hybrid Neural Network Watchdog For Outlier Detection." arXiv preprint arXiv:2103.00582 (2021).
	
	\bibitem{Chalom} Chalom, Edmond, Eran Asa, and Elior Biton. "Measuring image similarity: an overview of some useful applications." IEEE Instrumentation \& Measurement Magazine 16.1 (2013): 24-28.
	
	\bibitem{Chen} Chen, Chao, Jamie Twycross, and Jonathan M. Garibaldi. "A new accuracy measure based on bounded relative error for time series forecasting." PloS one 12.3 (2017): e0174202.
	
	\bibitem{Cohn} Cohn, D.,  L.E. Atlas, R. Ladner, M.A. El-Sharkawi, R.J. Marks II, M.E. Aggoune, D.C.
	Park ``Training connectionist networks with queries and selective sampling,'' Advances
	in Neural Network Information Processing Systems 2, Morgan Kaufman Publishers,
	Inc., San Mateo, CA. 1990.
	
	\bibitem{Creswell} Creswell, Antonia, et al. "Generative adversarial networks: An overview." IEEE Signal Processing Magazine 35.1 (2018): 53-65.
	
	\bibitem{Duren} Duren, Russell W., Robert J. Marks II, Paul D. Reynolds and Matthew L. Trumbo
	``Real-Time Neural Network Inversion on the SRC-6e Reconfigurable Computer,'' IEEE
	Transactions on Neural Networks, vol. 18, no. 3, May 2007 pp. 889-901.
	
	\bibitem{El-Sharkawi} El-Sharkawi, M.A., R.J. Marks II, Robert J. Streifel and I. Kerszenbaum ``Detection and Localization of Shorted-Turns in the DC-Field Winding of Turbine-Generator Rotors Using Novelty Filters and Fuzzified Neural Networks,’’ in Fuzzy System Theory in Electrical Power Engineering, M.E. El-Hawary, editor (IEEE Press, 1998), pp.85-111.
	
	\bibitem{Ferwerda} Ferwerda, James A., and Fabio Pellacini. "Functional difference predictors (FDPs): measuring meaningful image differences." ASILOMAR CONFERENCE ON SIGNALS SYSTEMS AND COMPUTERS. Vol. 2. IEEE; 1998, 2003.
	
	\bibitem{Fredieu} Fredieu, C. Tanner, et al. "Classification of Common Waveforms Including a Watchdog for Unknown Signals." arXiv preprint arXiv:2108.07339 (2021).
	
	\bibitem{Garg} Garg, Arunim, and Vijay Mago. "Role of machine learning in medical research: A survey." Computer Science Review 40 (2021): 100370.
	
	\bibitem{Golan} Golan, Izhak, and Ran El-Yaniv. "Deep anomaly detection using geometric transformations." arXiv preprint arXiv:1805.10917 (2018).
	
	\bibitem{Gondara} Gondara, Lovedeep. "Medical image denoising using convolutional denoising autoencoders." 2016 IEEE 16th International Conference on Data Mining Workshops (ICDMW). IEEE, 2016.
	
	\bibitem{Goodfellow} Goodfellow, Ian, et al. "Generative adversarial nets." Advances in neural information processing systems 27 (2014).
	
	\bibitem{Goodfellow2} Goodfellow, Ian, et al. "Generative adversarial networks." Communications of the ACM 63.11 (2020): 139-144.
	
	\bibitem{Guttormsson} Guttormsson, S., R.J. Marks II, M.A. El-Sharkawi and I. Kerszenbaum ``Elliptical novelty grouping for on-line short-turn detection of excited running rotors,’’ IEEE Transactions on Energy Conversion, IEEE Transactions on Volume: 14 1 , March 1999, pp. 16 -22.
	
	\bibitem{Hanga} Hanga, Khadijah M., and Yevgeniya Kovalchuk. "Machine learning and multi-agent systems in oil and gas industry applications: A survey." Computer Science Review 34 (2019): 100191.
	
	\bibitem{Haug} Haug, Samuel, Robert J. Marks, and William A. Dembski. "Exponential Contingency Explosion: Implications for Artificial General Intelligence." IEEE Transactions on Systems, Man, and Cybernetics: Systems (2021).
	
	\bibitem{Hendrycks} Hendrycks, Dan, Mantas Mazeika, and Thomas Dietterich. "Deep anomaly detection with outlier exposure." arXiv preprint arXiv:1812.04606 (2018).
	
	\bibitem{Hwang1990A} Hwang, J.N.  C.H. Chan, R.J. Marks II ``Frequency selective surface design based on
	iterative inversion of neural networks,'' Proceedings of the International Joint Conference
	on Neural Networks, San Diego, 17-21 June 1990, vol. I, pp.I39-I44.
	
	\bibitem{Hwang1990B} Hwang, J.N.,  J.J. Choi, S. Oh, R.J. Marks II ``Query learning based on boundary
	search and gradient computation of trained multilayer perceptrons,'' Proceedings of
	the International Joint Conference on Neural Networks, San Diego, June, 1990, 17-21
	June 1990, vol. III, pp.III57-III62.
	
	\bibitem{Hwang1990A} Hwang, J.N.  C.H. Chan, R.J. Marks II ``Frequency selective surface design based on
	iterative inversion of neural networks,'' Proceedings of the International Joint Conference
	on Neural Networks, San Diego, 17-21 June 1990, vol. I, pp.I39-I44.
	
	\bibitem{Hwang1990B} Hwang, J.N.,  J.J. Choi, S. Oh, R.J. Marks II ``Query learning based on boundary
	search and gradient computation of trained multilayer perceptrons,'' Proceedings of
	the International Joint Conference on Neural Networks, San Diego, June, 1990, 17-21
	June 1990, vol. III, pp.III57-III62.
	
	\bibitem{Jensen1997} Jensen, Craig A., Russell D. Reed, Mohamed A. El-Sharkawi, Robert J. Marks II
	``Location of Operating Points on the Dynamic Security Border Using Constrained
	Neural Network Inversion,'' Proceedings of the International Conference on Intelligent
	Systems Applications to Power Systems (ISAP), pp.209-217, Seoul, Korea, July 6-10,
	1997.
	
	\bibitem{Jensen1999A}  Jensen, Craig A., Reed, R.D.; Marks, R.J., II; El-Sharkawi, M.A.; Jae-Byung Jung;
	Miyamoto, R.T.; Anderson, G.M.; Eggen, C.J. ``Inversion of feedforward neural networks:
	algorithms and applications,'' Proceedings of the IEEE, Volume: 87 9, Sept.
	1999, Page(s): 1536 -1549
	
	\bibitem{Jensen1999B}  Jensen, Craig A., M.A. El-Sharkawi and R.J. Marks II ``Power Security Boundary Enhancement
	Using Evolutionary-Based Query Learning,'' Engineering Intelligent Systems,
	vol.7, no.9, pp.215-218 (December 1999). 
	
	\bibitem{Johnson} Johnson, Garrett M., and Mark D. Fairchild. "Measuring images: Differences, quality, and appearance." Human vision and electronic imaging VIII. Vol. 5007. International Society for Optics and Photonics, 2003.
	
	\bibitem{Karras} Karras, Tero, Samuli Laine, and Timo Aila. "A style-based generator architecture for generative adversarial networks." Proceedings of the IEEE/CVF Conference on Computer Vision and Pattern Recognition. 2019.
	
	\bibitem{Kassabalidis} Kassabalidis, I.N., Mohamed El-Sharkawi, Robert J. Marks II “Border Identification
	For Power System Security Assessment Using Neural Network Inversion: An Overview,”
	2002 Congress on Evolutionary Computation, 2002 IEEE World Congress on Computational
	Intelligence May 12-17, 2002, Honolulu, pp.1075-1079.
	
	\bibitem{Kim} Kim, Juhwan, Seokyong Song, and Son-Cheol Yu. "Denoising auto-encoder based image enhancement for high resolution sonar image." 2017 IEEE Underwater Technology (UT). IEEE, 2017.
	
	\bibitem{Lore} Lore, Kin Gwn, Adedotun Akintayo, and Soumik Sarkar. "LLNet: A deep autoencoder approach to natural low-light image enhancement." Pattern Recognition 61 (2017): 650-662.
	
	\bibitem{Lu} Lu, Xugang, et al. "Speech enhancement based on deep denoising autoencoder." Interspeech. 2013.
	
	\bibitem{Mao} Mao, Xudong, et al. "Least squares generative adversarial networks." Proceedings of the IEEE international conference on computer vision. 2017.
	
	\bibitem{Marchi} Marchi, Erik, et al. "A novel approach for automatic acoustic novelty detection using a denoising autoencoder with bidirectional LSTM neural networks." 2015 IEEE international conference on acoustics, speech and signal processing (ICASSP). IEEE, 2015.
	
	\bibitem{Metz} Metz, Luke, et al. "Unrolled generative adversarial networks." arXiv preprint arXiv:1611.02163 (2016).

	\bibitem{Miko} Mikołajczyk, Agnieszka, and Michał Grochowski. "Data augmentation for improving deep learning in image classification problem." 2018 international interdisciplinary PhD workshop (IIPhDW). IEEE, 2018.
	
	\bibitem{Narayanan}  Narayanan, Sreeram, R.J. Marks II , John L. Vian, J.J. Choi, M.A. El-Sharkawi and
	Benjamin B. Thompson ``Set Constraint Discovery: Missing Sensor Data Restoration
	Using Auto-Associative Regression Machines,’’ Proceedings of the 2002 International
	Joint Conference on Neural Networks, 2002 IEEE World Congress on Computational
	Intelligence, May12-17, 2002, Honolulu, pp. 2872-2877.
	
	\bibitem{Nguyen} Nguyen, Duc Tam, et al. "Anomaly detection with multiple-hypotheses predictions." International Conference on Machine Learning. PMLR, 2019.
	
	\bibitem{Oh}  Oh, Seho,  R.J. Marks II and M.A. El-Sharkawi ``Query based learning in a multilayered
	perceptron in the presence of data jitter,'' Applications of Neural Networks to Power
	Systems, (Proceedings of the First International Forum on Applications of Neural
	Networks to Power Systems), July 23-26, 1991, Seattle, WA, (IEEE Press, pp.72-75).
	
	
	\bibitem{Ozbayoglu} Ozbayoglu, Ahmet Murat, Mehmet Ugur Gudelek, and Omer Berat Sezer. "Deep learning for financial applications: A survey." Applied Soft Computing 93 (2020): 106384.
	
	\bibitem{Paleyes} Paleyes, Andrei, Raoul-Gabriel Urma, and Neil D. Lawrence. "Challenges in deploying machine learning: a survey of case studies." arXiv preprint arXiv:2011.09926 (2020).
	
	\bibitem{Park} Park, Seonhee, et al. "Dual autoencoder network for retinex-based low-light image enhancement." IEEE Access 6 (2018): 22084-22093.
	
	\bibitem{Paszke} Paszke, Adam, et al. "Pytorch: An imperative style, high-performance deep learning library." Advances in neural information processing systems 32 (2019): 8026-8037.

	\bibitem{Perez} Perez, Luis, and Jason Wang. "The effectiveness of data augmentation in image classification using deep learning." arXiv preprint arXiv:1712.04621 (2017).
	
	\bibitem{Pradhan} Pradhan, Manoranjan, Sateesh Kumar Pradhan, and Sudhir Kumar Sahu. "Anomaly detection using artificial neural network." International Journal of Engineering Sciences \& Emerging Technologies 2.1 (2012): 29-36.
	
	\bibitem{Qi} Qi, Yumei, et al. "Stacked sparse autoencoder-based deep network for fault diagnosis of rotating machinery." Ieee Access 5 (2017): 15066-15079.
	
	\bibitem{Reed} Reed, Russell D. and Robert J. Marks II ``An Evolutionary Algorithm for Function
	Inversion and Boundary Marking,'' Proceedings of the IEEE International Conference
	on Evolutionary Computation, p. 794-797, Perth, Australia. November 26-30, 1995.
	
	\bibitem{Rosin} Rosin, Paul L., and Tim J. Ellis. "Image difference threshold strategies and shadow detection." BMVC. Vol. 95. 1995.
	
	\bibitem{Rabe} Rabe, Martin, Stefan Milz, and Patrick Mader. "Development Methodologies for Safety Critical Machine Learning Applications in the Automotive Domain: A Survey." Proceedings of the IEEE/CVF Conference on Computer Vision and Pattern Recognition. 2021.
	
	\bibitem{Sabokrou} Sabokrou, Mohammad, et al. "Deep-anomaly: Fully convolutional neural network for fast anomaly detection in crowded scenes." Computer Vision and Image Understanding 172 (2018): 88-97.
	
	\bibitem{Sakurada} Sakurada, Mayu, and Takehisa Yairi. "Anomaly detection using autoencoders with nonlinear dimensionality reduction." Proceedings of the MLSDA 2014 2nd workshop on machine learning for sensory data analysis. 2014.
	
	\bibitem{Sharp} Sharp, Michael, Ronay Ak, and Thomas Hedberg Jr. "A survey of the advancing use and development of machine learning in smart manufacturing." Journal of manufacturing systems 48 (2018): 170-179.
	
	\bibitem{Shorten} Shorten, Connor, and Taghi M. Khoshgoftaar. "A survey on image data augmentation for deep learning." Journal of Big Data 6.1 (2019): 1-48.
	
	\bibitem{Stallkamp} Stallkamp, Johannes, et al. "The German traffic sign recognition benchmark: a multi-class classification competition." The 2011 international joint conference on neural networks. IEEE, 2011.
	
	\bibitem{Streifel} Streifel, R.J., R.J. Marks II, M.A. El-Sharkawi and I. Kerszenbaum ``Detection of
	Shorted-Turns in the Field Winding of Turbine-Generator Rotors Using Novelty Detectors:
	Development and Field Test,’’ IEEE Transactions on Energy Conversion, vol.11,
	no.2, June 1996, pp.312-317.
	
	\bibitem{Taylor} Taylor, Luke, and Geoff Nitschke. "Improving deep learning with generic data augmentation." 2018 IEEE Symposium Series on Computational Intelligence (SSCI). IEEE, 2018.
	
	\bibitem{Thompson2002}  Thompson, Benjamin B., Robert J Marks II , Jai J Choi, Mohamed A El-Sharkawi
	``Implicit Learning in Autoencoder Novelty Assessment,’’ Proceedings of the 2002 International
	Joint Conference on Neural Networks, 2002 IEEE World Congress on Computational
	Intelligence, May12-17, 2002, Honolulu, pp. 2878-2883.
	
	\bibitem{Thompson2003}  Thompson, Benjamin B., Robert J. Marks II, and Mohamed A. El-Sharkawi ``On the Contractive Nature of Autoencoders: Application to Missing Sensor Restoration,’’ 2003 International Joint Conference on Neural Networks, July 20-24, 2003 , Portland, Oregon (pp. 3011-3016)
	
	
	\bibitem{Tsang1991} Tsang, Leung, Z. Chen, S. Oh, R.J. Marks II and A.T.C. Chang ``Inversion of snow parameters
	from passive microwave remote sensing measurements by a neural network trained
	with a multiple scattering model,'' Proceedings of the 1991 International Geoscience
	and Remote Sensing Symposium, 3-7 June 1991, Espoo, Finland.
	
	\bibitem{Tsang1992} Tsang, Leung, Z. Chen, S. Oh, R.J. Marks II and A.T.C. Chang ``Inversion of snow parameters
	from passive microwave remote sensing measurements by a neural network
	trained with a multiple scattering model,'' IEEE Transactions on Goescience and Remote
	Sensing, vol. 30, no.5, pp. 1015-1024 (1992).
	
	\bibitem{Wang} Wang, Zhou, and Alan C. Bovik. "Mean squared error: Love it or leave it? A new look at signal fidelity measures." IEEE signal processing magazine 26.1 (2009): 98-117.
	
	\bibitem{Wang2} Z. Wang, A. C. Bovik, H. R. Sheikh and E. P. Simoncelli, “Image quality assessment: From error visibility to structural similarity,” IEEE Transactions on Image Processing, vol. 13, no. 4, pp. 600-612, Apr. 2004.
	
	\bibitem{Wei} Wei, Qi, et al. "Anomaly detection for medical images based on a one-class classification." Medical Imaging 2018: Computer-Aided Diagnosis. Vol. 10575. International Society for Optics and Photonics, 2018.
	
	\bibitem{Wen} Wen, Qingsong, et al. "Time series data augmentation for deep learning: A survey." arXiv preprint arXiv:2002.12478 (2020).
	
	\bibitem{Zhou} Zhou, Chong, and Randy C. Paffenroth. "Anomaly detection with robust deep autoencoders." Proceedings of the 23rd ACM SIGKDD international conference on knowledge discovery and data mining. 2017.
	
	\bibitem{Zhu} Zhu, Xinyue, et al. "Data augmentation in emotion classification using generative adversarial networks." arXiv preprint arXiv:1711.00648 (2017).
	
\end{thebibliography}
\end{document}